\documentclass[a4paper]{./styles/svproc}
\usepackage{url}
\usepackage{glossaries}
\makeglossaries
\usepackage{bbm}
\usepackage{amsfonts}
\usepackage{amsmath}
\usepackage{graphicx}
\usepackage{pdfpages}
\usepackage{multirow}
\usepackage{float} % for fixing position of figures
\graphicspath{{figures/}}
\newcommand{\bld}{\mathbf}

\newcommand{\RNum}[1]{\uppercase\expandafter{\romannumeral #1\relax}}

\usepackage{color}
\newcommand{\fref}[1]{Fig.~\ref{#1}}
\newcommand{\sref}[1]{Section~\ref{#1}}
\newcommand{\ssref}[1]{subsection~\ref{#1}}

\newcommand{\tref}[1]{Table~\ref{#1}}

\begin{document}
%keep this alphapetically ordered
% \newglossaryentry{BNN}{name=BNN, description={Bayesian Neural Network},first={\glsentrydesc{BNN} (\glsentrytext{BNN}) },plural={BNNs},descriptionplural={Bayesian Neural Networks},firstplural={\glsentrydescplural{BNN} (\glsentryplural{BNN})}}
\newglossaryentry{BNN}{name=BNN, description={Bayesian Neural Network},first={Bayesian Neural Network (BNN)}}

\newglossaryentry{CDP}{name=CDP, description={Concrete Dropout},first={Concrete Dropout (CDP)}}

% \newglossaryentry{CNN}{name=CNN, description={Convolutional Neural Network},first={\glsentrydesc{CNN} (\glsentrytext{CNN})}, plural={CNNs},descriptionplural={Convolutional Neural Networks},firstplural={\glsentrydescplural{CNN} (\glsentryplural{CNN})}}
\newglossaryentry{CNN}{name=CNN, description={Convolutional Neural Network},first={Convolutional Neural Network (CNN)}}

\newglossaryentry{CRF}{name=CRF, description={Conditional Random Field},first={Conditional Random Field (CRF)}}

% \newglossaryentry{DNN}{name=DNN, description={Deterministic Neural Network},first={\glsentrydesc{DNN} (\glsentrytext{DNN})}, plural={DNNs},descriptionplural={Deterministic Neural Networks},firstplural={\glsentrydescplural{DNN} (\glsentryplural{DNN})}}
\newglossaryentry{DNN}{name=DNN, description={Deterministic Neural Network},first={Deterministic Neural Network (DNN)}}
	
\newglossaryentry{ECE}{name=ECE, description={Expected Calibration Error},first={Expected Calibration Error (ECE)}}
\newglossaryentry{ELBO}{name=ELBO, description={Evidence Lower Bound},first={Evidence Lower Bound (ELBO)}}
\newglossaryentry{KL-div}{name=KL-div, description={Kullbach-Leibler divergence},first={Kullbach-Leibler divergence (KL-div)}}
\newglossaryentry{LAP}{name=LAP, description={Laplace Approximation},first={Laplace Approximation (LAP)}}
\newglossaryentry{LBP}{name=LBP, description={Loopy Belief Propagation},first={Loopy Belief Propagation (LBP)}}
\newglossaryentry{MAP}{name=MAP, description={Maximum a posterior },first={Maximum a posterior (MAP)}}
\newglossaryentry{MCD}{name=MCD, description={Monte Carlo Dropout},first={Monte Carlo Dropout (MCD)}}
\newglossaryentry{MCE}{name=MCE, description={Maximal Calibration Error},first={Maximal Calibration Error (MCE)}}
\newglossaryentry{MCMC}{name=MCMC, description={Markov Chain Monte Carlo},first={Markov Chain Monte Carlo (MCMC)}}
\newglossaryentry{MLE}{name=MLE, description={Maximum Likelihood Estimation},first={Maximum Likelhood Esitmation (MLE)}}
\newglossaryentry{MLP}{name=MLP, description={Multi-layer Perceptron},first={Multi-layer Perceptron (MLP)}}
\newglossaryentry{NLL}{name=NLL, description={Negative Log Likelihood},first={Negative Log Likelihood (NLL)}}
\newglossaryentry{OOD}{name=OOD, description={Out-of-distribution},first={Out-of-distribution (OOD)}}
\newglossaryentry{PR}{name=PR, description={Precision Recall},first={Precision Recall (PR)}}

% \newglossaryentry{PGM}{name=PGM, description={Probabilistic Graphical Model},first={\glsentrydesc{PGM}~(\glsentrytext{PGM})}, plural={PGMs},descriptionplural={Probabilistic Graphical Models},firstplural={\glsentrydescplural{PGM}~(\glsentryplural{PGM})}} 
\newglossaryentry{PGM}{name=PGM, description={Probabilistic Graphical Model},first={Probabilistic Graphical Model (PGM)}}
\newglossaryentry{ROC}{name=ROC, description={Receiver Operating Characteristic},first={Receiver Operating Characteristic (ROC)}}
\newglossaryentry{SGD}{name=SGD, description={Stochastic Gradient Descent},first={Stochastic Gradient Descent (SGD)}}
\newglossaryentry{SGLD}{name=SGLD, description={Stochastic Gradient Langevin Dynamics},first={Stochastic Gradient Langevin Dynamics (SGLD)}}
%\newglossaryentry{R.V.}{name=R.V., description={Random Variable},first={Random Variable (R.V.)}}
\newglossaryentry{VI}{name=VI, description={Variational Inference},first={Variational Inference (VI)}}
\newglossaryentry{WRGBD}{name=WRGB-D, description={RGB-D Dataset from Washington University~\cite{lai2011large}},first={RGB-D Dataset from Washington University (WRGB-D)~\cite{lai2011large}}}

\mainmatter              % start of a contribution
%
% \title{Bayesian Neural Nets for Uncertainty-based Improvement in Adaptive Classification}
%\title{Uncertainty-based Improvements via Smoothed Predictions from Bayesian Neural Networks}
\title{Introspective Robot Perception using Smoothed Predictions from Bayesian Neural Networks}
\titlerunning{Introspective Perception using BNNs}  % abbreviated title (for running head)
%                                     also used for the TOC unless
%                                     \toctitle is used
%
\author{Jianxiang Feng$^{*1}$ \and  Maximilian Durner$^{*1}$ \and \\ Zolt\'an-Csaba M\'arton$^1$ \and Ferenc B\'alint-Bencz\'edi$^2$ \and Rudolph Triebel$^{1,3}$}
\authorrunning{Feng et al.} % abbreviated author list (for running head)
%
%%%% list of authors for the TOC (use if author list has to be modified)
% \tocauthor{Ivar Ekeland, Roger Temam, Jeffrey Dean, David Grove,
% Craig Chambers, Kim B. Bruce, and Elisa Bertino}
%
\institute{$^*$ Equal contribution;
$^1$ Institute of Robotics and Mechatronics, German Aerospace Center (DLR),
\email{first.last@dlr.de};
$^2$ Institute for Artificial Intelligence, University of Bremen,
\email{balintbe@cs.uni-bremen.de};
$^3$ Department of Computer Science, Technical University of Munich,
\email{rudolph.triebel@in.tum.de}}

\maketitle              % typeset the title of the contribution

\begin{abstract}
%We present a novel approach to classify objects from RGB images, where we mainly focus on predictions that also provide a reliable uncertainty estimate.
%To achieve this we employ a \gls{BNN}, and we consider both currently used strategies to obtain uncertainty, namely Monte-Carlo dropout and the Kronecker-factored Laplace Approximation.
%Furthermore, we incorporate contextual information using a co-occurence matrix of object instances and use that to model pair-wise potentials of a \gls{CRF}.
%Finally, we exploit these uncertainties to provide an efficient pipeline for learning class labels adaptively.
%We show the higher effectiveness of our approach compared to standard approaches in experiments on benchmark data sets that are particularly relevant for robot perception tasks.  
This work focuses on improving uncertainty estimation in the field of object classification from RGB images and demonstrates its benefits in two robotic applications. %, since this is still a missing feature for robotic deployment.
We employ a \gls{BNN}, and evaluate two practical inference techniques to obtain better uncertainty estimates, namely \gls{CDP} and Kronecker-factored \gls{LAP}.
%The reliable uncertainty estimates are applied in two robotic applications resulting in increased performance values.
We show a performance increase using more reliable uncertainty estimates as unary potentials within a \gls{CRF}, which is able to incorporate contextual information as well.
% Furthermore, the obtained uncertainties are exploited to provide an efficient pipeline for learning class labels adaptively.
Furthermore, the obtained uncertainties are exploited to achieve domain adaptation in a semi-supervised manner, which requires less manual efforts in annotating data.
We evaluate our approach on two public benchmark datasets that are relevant for robot perception tasks. %, and we also recorded extensions to these that will be made public.

%The abstract should summarize the contents of the paper
%using at least 70 and at most 150 words. It will be set in 9-point
%font size and be inset 1.0 cm from the right and left margins.
%There will be two blank lines before and after the Abstract. \dots
, introspective classification
\end{abstract}
\section{Introduction}
Visual scene understanding plays an important role in the field of robotic perception.
In recent years, deep learning showed promising results within this context (e.g. object classification, detection or segmentation).
%In recent years, deep learning has successfully been applied to several robotics related vision tasks such as object classification, detection or segmentation.
Yet, although the applied deep neural networks outperform most traditional methods, they lack 
%-- due to the characteristic of usually applied softmax -- %
a significant property for robots in real world: a reliable uncertainty estimation.
Advanced robotics highly rely on perceptual systems in order to be able to understand and adapt to its environment.
Providing also the confidence of predictions based on the perceived information enhances the ability of robotic systems even further.
%Supplying the robotic system  of uncertainty for such (visual) predictive systems 
It equips robots with the ability to know when it does and when it does not know.
%More than that, corresponding counter measures can be taken to avoid unnecessary accidents.
%Concrete examples include the steering control in self-driving cars~\cite{mcallister2017concrete}, disease diagnosis in the medical domain~\cite{leibig2017leveraging}, or even applications in aerospace or other domains requiring higher precision and stronger robustness. 
%Besides the safety issue~\cite{amodei2016concrete} -- for the robot itself and its surroundings -- introspection about the predictions also has a positive impact on decision making, failure recovery and human-robot interaction.
Besides the safety issue -- for the robot itself and its surroundings -- introspection about the predictions also has a positive impact on decision making, failure recovery and human-robot interaction.
% A robot which is unsure about the object to grasp could decide to have a second view on the object.
% Another important field is the human-robot interaction.
% Uncertainty can be treated as a bridge between machine learning algorithms and humans.
% Such interaction or the exploration of the environment can establish a more robust and more data-efficient machine learning system.
%The idea behind that is, with reliable uncertainty estimation we know which data sample the model is unfamiliar with and thus providing more information for the model.
% The uncertainty estimation indicates when the model requires help from human experts.
%For instance, in industrial component inspection task, firstly a model which can perform quite well on some evaluation datasets is trained. With the assumption that all predictions made by the model are correct, the system is deployed to the real application.
%However, there is no guarantee that this model represents the real distribution of data perfectly and is robust against a slight or even large domain gap.
%If the distributions of the real world data vary a lot because of unexpected factors e.g. equipment aging or changes of weather condition, the model could fail silently and lead to unexpected accidents.
%With reliable model uncertainty estimation, this kind of weakness can be mitigated or even eliminated. 
Furthermore, reliable uncertainty estimation is beneficial for active learning~\cite{gal2017deep}, reinforcement learning~\cite{blundell2015weight,osband2016deep,gal2016improving}, detection of the unknown classes and adversarial attacks~\cite{grimmett15introspective,hendrycks2016baseline,kurakin2016adversarial}.
Recent research on improving the uncertainty estimation of deep neural networks includes \gls{BNN}s\cite{blundell2015weight,balan2015bayesian,gal2016dropout,louizos2016structured,gal2017concrete,sun2017learning,louizos2017multiplicative,ritter2018scalable,wang2018adversarial}, bootstrapping~\cite{osband2016deep}, ensemble methods~\cite{lakshminarayanan2017simple} and so on. 
Among them, a \gls{BNN} is more theoretically sound and able to provide promising performances. 
By taking into account the practicality in real-world applications, we evaluate \gls{BNN}s with two inference techniques which are \gls{CDP}~\cite{gal2017concrete} and \gls{LAP}~\cite{ritter2018scalable} in term of comprehensive metrics. 
However, we are more curious about the question, to which extent the improved uncertainty estimates can boost the performances on uncertainty-relevant tasks.
Therefore, in this work we focus on studying the improvements by exploiting uncertainty estimates from \gls{BNN}s which are demonstrated by applying them to \textbf{(1)} support \gls{CRF}s which can incorporate additional contextual information as well and \textbf{(2)} reduce the manual efforts for data annotations in domain adaptation tasks.

\begin{figure}[]	
	\begin{center}
		\includegraphics[width=\columnwidth, height=4.9cm]{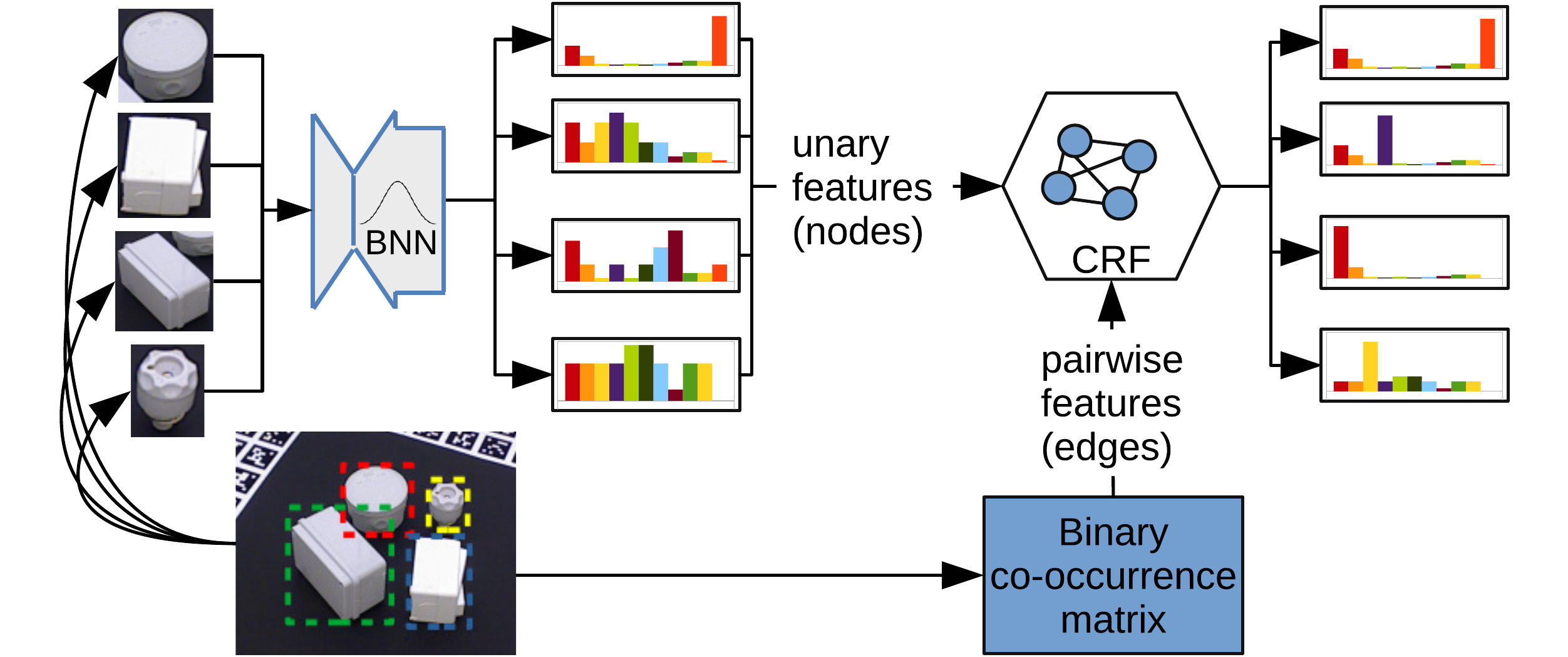}
		\caption{The combination of \gls{BNN}s and \gls{CRF}s: the predictive distributions of objects in the scene from \gls{BNN}s serve as unary features in the \gls{CRF}s, which can take into account the contextual information from the scene of objects.}		
		\label{fig:combined_crf}
	\end{center}
\end{figure}

In the line of combining deep learning and \gls{PGM}~\cite{koller2009probabilistic}, previous works~\cite{tompson2014joint,liu2015deep,wang2016towards,johnson2016composing} mainly focus on joint training of these two kinds of model in order to share the advantages of both, which are  abilities of expressive representation learning and structured learning, respectively. 
None of them emphasize the role of uncertainty estimation when combining them as sub-modules, which can improve the robustness of the system in practical applications such as real world robotics.
% On the one hand, reliable uncertainty estimates contain more information than a single prediction from traditional \glspl{DNN}, which are shown to be overconfident~\cite{guo2017calibration}. % \com{do we need DNN and DNN?!}
%This kind of information should be utilized for further improvements. 
%Since the uncertainty is modeled in a probabilistic manner, other \glspl{PGM} are an ideal choice for this purpose.
% According to the probabilistic characteristic of uncertainty captured via \gls{BNN}, other \glspl{PGM} are an ideal choice for this purpose. 
%Nevertheless, we observe that there are rich contextual information in scene object recognition that can help to disambiguate objects with similar appearances but different contexts. 
%Therefore, in order to exploit better uncertainty estimation and contextual information, we show that a \gls{CRF} can achieve further improvements in this way on a challenging dataset.
In this work, we propose to use uncertainty estimates to improve classification by combining \gls{CRF}s (see \fref{fig:combined_crf}).

On the other hand, robots deployed in a new situation are often confronted with environmental changes and novel objects. 
Nevertheless, in most of the time a base classifier trained on an easily obtainable dataset (e.g. public large-scale or synthetic) is available beforehand. 
The classifier needs to be adapted to the test environment, while the manual efforts of collecting and annotating the adaptation data should be kept as low as possible.
This requirement can be cast into the field of domain adaptation in a self/semi-supervised manner. 
Self-supervised learning refers to learning with self-provided supervisions such as geometrical cues within images~\cite{kolesnikov2019revisiting} instead of strong but laborious human-supervisions and these self-supervisions can be extended to self-generated pseudo labels by the model itself, which can be used for domain adaptation naturally~\cite{tang2012shifting,zou2018unsupervised,xu2019self}. 
This task can also be framed into a semi-supervised manner, when a small amount of manual annotations are allowed to be taken into the procedure~\cite{wang2016cost,lin2017active}.
Among these prior works, none of them highlights the importance of uncertainty estimates which can help distinguishing true positives (served for automatic-annotation) and false positives in both self-supervised and semi-supervised manner. 
As far as we know, we are the first to make use of uncertainty estimates from \gls{BNN}s in this kind of tasks.

The remaining of the paper is organized as follows:
we review prior works in the related areas in section 2.
While section 3 recaps the theoretical concept of \gls{BNN}s,
section 4 explains our proposed approaches. Then we show experimental results demonstrating their effectiveness in section 5 and conclude in section 6. 

\section{Related Work}
A \textbf{\gls{BNN}} 
%Regarding model uncertainty or quality measure that acts similar to model uncertainty, there are some other more theoretical approaches which employ \gls{BNN}~\cite{mackay1992practical,neal2012bayesian}, bootstrapping~\cite{osband2016deep}, ensemble methods~\cite{lakshminarayanan2017simple} or even ensemble of \gls{BNN}~\cite{smith2018understanding}. 
\cite{mackay1992practical,neal2012bayesian} provides a principal way to obtain model uncertainty by considering the distribution on model parameters. However, it has difficulty scaling to complex network architectures and large training sets nowadays.
%In \gls{VI}, this inference problem is cast into an optimization problem by minimizing the \gls{KL-div} between the approximate posterior distribution and the real posterior. Graves~\cite{graves2011practical} firstly employed a fully factorized Gaussian with a biased gradient estimator on a practical problem. 
Besides sampling based methods \cite{balan2015bayesian,wang2018adversarial}, \gls{VI} \cite{graves2011practical} suits practical applications due to its ability of fast inference. 
In the era of deep learning, there is a bunch of research works in this direction \cite{blundell2015weight,louizos2016structured,sun2017learning,louizos2017multiplicative,ritter2018scalable}. \gls{CDP} \cite{gal2017concrete} is an extension of \gls{MCD}~\cite{gal2016dropout} which can learn dropout rates from the data without efforts of manual tuning. More than that, \gls{CDP} can be inserted into existing network architectures very easily. On the other hand \gls{LAP} does not require re-training and thus suits most of the already-trained networks as well.

\textbf{Combination of deep learning and \gls{PGM}s:} Liu et al.~\cite{liu2015deep} trained a \gls{CNN} and \gls{CRF} jointly for depth estimation, while Tompson et al.~\cite{tompson2014joint} integrated Markov random fields with \gls{CNN} for pose estimation. 
Wang et al.~\cite{wang2016towards} combined deep learning with Bayesian networks for recommendation systems and topic models. 
Johnson et al.~\cite{johnson2016composing} proposed Structured variational autoencoder (SVAE) to learn a structured and thus more interpretable latent representation.
Our work differ from them in the way of training. 
Since we want to evaluate the effects of uncertainty estimates, it's better to analyze them separately. Similar to us, Liu et al. \cite{liu2015crf} combined features learned from deep neural nets and \gls{CRF} for segmentation tasks. But they trained another classifier with these features for the unary potentials without evaluating the effects of uncertainty estimates.
% Nevertheless, to investigate the effect of jointly training these two kinds of model with better uncertainty estimates is an interesting research direction in the future.   

% The correlation between objects in specific scene are of importance and can be represented by a co-occurrence matrix~\cite{galleguillos2008object}.

\textbf{Semi/Self-supervised domain adaptation:} Some works~\cite{kolesnikov2019revisiting,xu2019self} aim to learn a more generalized feature distribution via designing specific \textit{pretext} tasks without explicit human supervisions (e.g. class labels). 
Others \cite{lin2017active,wang2016cost,zou2018unsupervised} tried to employ true positives as self-supervisions for adaptation. Zou et al.~\cite{zou2018unsupervised} mentioned the class imbalance problem and proposed to mitigate it by normalizing the class-wise confidence. To note that this problem is obvious in this kind of task, which was verified and mitigated by class-balanced augmentations in our experiments.

% One main difference from them is that we only demonstrate the improvements brought by better uncertainty estimation instead of formulating the problem as continuous and incremental learning and running experiments . 
% However, our work can be adapted into continuous learning easily.

\section{Bayesian Neural Networks}
% In this section, we review the concepts that form the basis of the method we propose. 
% We start with the introduction of the \gls{BNN} and its inference techniques, which are used for obtaining reliable uncertainty estimate. 
% Next the \gls{CRF} is introduced, which can make use of better uncertainty estimation and incorporates contextual information for further improvement. Unless otherwise stated, we use boldfaced uppercase and lowercase letters to denote matrices and column vectors respectively.

In general, a neural network can be modelled as a function $f^{\boldsymbol{\omega}}(\mathbf{x})=\mathbf{y}$ that maps from an input space $\mathcal{X}$ to an output space $\mathcal{Y}$, where $\boldsymbol{\omega}= \{W_{1:L}, \mathbf{b}_{1:L}\}$ are the weights of the network consisting of matrices $W_i$ and biases $\mathbf{b}_i$ for each of its $L$ layers. 
In the training phase, the weights $\boldsymbol{\omega}$ are determined by optimizing a loss function $E(f^{\boldsymbol{\omega}}(\mathbf{x}_i),\mathbf{y}_i)$ for a given training data set $\mathcal{D}=\{(\mathbf{x}_i,\mathbf{y}_i)_{i=1}^N\}$. 
In contrast, a Bayesian Neural Network (\gls{BNN}) not only aims to find an optimal $\boldsymbol{\omega}$, but also defines a \emph{posterior distribution} $p(\boldsymbol{\omega}\mid \mathcal{D})$. %, where $X$ and $Y$ are the matrices consisting of all training data samples.  
Given this posterior, inference on a new test sample $(\mathbf{x}^\ast, \mathbf{y}^\ast)$ can be done using the \emph{predictive distribution}
\begin{equation}
\label{eq:pred_dist}
  p(\mathbf{y}^\ast\mid \mathbf{x}^\ast, \mathcal{D} ) = 
\int p(\mathbf{y}^\ast\mid \mathbf{x}^\ast,\boldsymbol{\omega}) p(\boldsymbol{\omega} \mid \mathcal{D} ) d \boldsymbol{\omega},
\end{equation}
where for classification tasks the likelihood $p(\mathbf{y}^\ast\mid \mathbf{x}^\ast,\boldsymbol{\omega})$ is usually obtained from the \emph{softmax} of the prediction  $f^{\boldsymbol{\omega}}(\mathbf{x}^\ast)$. The benefit of using \eqref{eq:pred_dist} for predictions instead of only using the likelihood is that the model also incorporates the \emph{epistemic} uncertainty, i.e. the one that stems from incorrect model parameters, thereby providing better (less overconfident) uncertainty estimates. 

Unfortunately, obtaining the parameter posterior $p(\boldsymbol{\omega}\mid \mathcal{D})$ is not tractable in all but the simplest cases due to the high dimensionality of the parameter space. %$\mathcal{W}$.  
Therefore, approximations need to be used, and we investigate two common ones: the \gls{CDP} and Kronecker-factored \gls{LAP}.

\subsection{Concrete dropout}
Dropout \cite{srivastava2014dropout} was originally proposed to regularize the training process of a \gls{DNN} to improve their generalization performance, although yet without a formal interpretation.  Then, Gal \cite{gal2016uncertainty} showed that using dropout can be interpreted as sampling from a distribution $q_\theta(\boldsymbol{\omega})$ that approximates the posterior $p(\boldsymbol{\omega}\mid X, Y)$ in terms of the KL-divergence
\begin{equation}
\label{eq:kldiv}
KL(q_{\theta}(\boldsymbol \omega)\|p(\bld \omega \mid \mathcal{D}))  = - \int q_{\theta}(\boldsymbol \omega) \log \frac{p(\boldsymbol \omega \mid \mathcal{D}) }{q_{\theta}(\boldsymbol \omega) }.
\end{equation}
where $\theta = \{\boldsymbol \omega,\bld p\}$, $\bld p$ is the vector of dropout rates of layers in which dropout is inserted. Minimizing this is equivalent to minimizing the \gls{ELBO} 
\begin{eqnarray}
\label{eq:kldiv}
\mathcal{L}({\theta}) &=& - \sum_{i=1}^N\int q_\theta(\boldsymbol \omega) \log p(\bld y_i \mid f^{\boldsymbol \omega}(\bld x_i) )d \boldsymbol \omega + \mbox{KL}( q_\theta(\boldsymbol \omega) || p(\boldsymbol \omega)) \\
&\approx&  - \sum_{i\in \mathcal{S}}\frac{N}{K}\int q_\theta(\boldsymbol \omega) \log p(\bld y_i \mid f^{\boldsymbol \omega}(\bld x_i) ) d \boldsymbol \omega  + \mbox{KL}( q_\theta(\boldsymbol \omega) || p(\boldsymbol \omega)),
\end{eqnarray}
where $\mathcal{S}$ is a mini-batch of size $K$. 
To estimate the expected log likelihood in the first term, Monte Carlo integration is used, i.e. samples are generated from $q_\theta(\boldsymbol \omega)$, and the integral is approximated by summing likelihood terms over the samples. 
The problem here is that using this standard method, this first term can not be derived with respect to $\theta$, which is necessary to minimize $\mathcal{L}({\theta})$. 
Therefore, the \emph{re-parameterization trick} is used, i.e. a bivariate transformation $g(\theta,\epsilon)$ is used to separate the parameters $\theta$ from samples $\epsilon \sim p(\epsilon)$ that are generated from a distribution with fixed parameters. 
Originally, this could be done only for a Gaussian dropout distribution, later Gal \emph{et al.} \cite{gal2017concrete} showed that for Bernoulli dropout, a \emph{con}tinuous relaxation of this dis\emph{crete} distribution can be found, i.e. a concrete distribution \cite{maddison2016concrete}, which can then be derived wrt. $\theta$ for optimization. 
This is denoted \emph{concrete dropout}. 
In our experiments, we use the implementation provided by Gal \emph{et al.} \cite{gal2017concrete}.

\subsection{Laplace approximation}
The idea within the so-called Laplace approximation is to employ a second-order Taylor expansion at the maximum of the log posterior:
%The second approach to approximate the parameter posterior that we consider is the Laplace approximation. Here, the idea is to employ a second-order Taylor expansion at the maximum of the log posterior:
\begin{equation}
  \log p(\bld\omega \mid  X, Y) \approx \log  p(\bld\omega^\ast \mid  X, Y) - \frac{1}{2}(\boldsymbol \omega - \boldsymbol \omega^\ast)^TH (\boldsymbol \omega - \boldsymbol \omega^\ast),
\end{equation}
where $\boldsymbol \omega^\ast$ is the parameter vector that maximizes the log posterior and $H$ is the Hessian of the negative log posterior. 
Note that the first derivative vanishes at $\boldsymbol \omega^\ast$ and $H$ is p.s.d. because $\boldsymbol \omega^\ast$ is assumed to be a local maximum. 
After taking the exponential and normalizing we obtain 

\begin{equation} 
\label{gaussian form}
p(\bld\omega \mid  X, Y) \approx \mathcal N(\boldsymbol \omega^\star,  H ^{-1}).
\end{equation}

 Unfortunately, the dimensionality of this multi-variate normal distribution is in most cases too high to be practical. 
 Also, $H$ needs to be computed on the entire data set, which is also infeasible. Instead, it is approximated by the expected Hessian $\mathbb{E}_{p(X, Y)} [H]$, computed on mini-batches.  
 To reduce the dimensionality, a first step is to assume independence across the layers of the DNN, i.e. $H$ is block-diagonal with $L$ blocks $H_i$, one for each layer. 
 
 % \subsubsection{Scalable Laplace approximation}
 %Besides evaluating dropout variational inference, another scalable deterministic approximate inference technique~\cite{ritter2018scalable} is considered for comparison. Laplace approximation is to impose a Gaussian distribution over the local optimizer to approximate the real posterior: 
 % \begin{equation} \label{gaussian form}
 % \boldsymbol \omega \sim \mathcal N(\boldsymbol \omega^\star, \bld H ^{-1})
 % \end{equation}
 % where $\boldsymbol \omega^\star$ is one local optimizer and covariance matrix is the inverse of negative Hessian $\bld H^{-1}$.
 % This local optimizer can be obtained in any \glspl{DNN} that have been trained. 
 % Consequently, this approach is appealing because it can be used for every \gls{DNN} without re-training the network. 
 % However, to estimate the covariance matrix of the approximate Gaussian is infeasible because of high storage overhead and computation complexity.
 % If the size of training set is too large, the computation of Hessian is infeasible, it can be approximated by the expected negative Hessian $\mathbb{E}_{p(\bld X, \bld Y)} [\bld H]$, which can be calculated in mini-batches feasibly. 
 Under certain conditions, the Fisher information matrix $F$, which is the outer product of the first derivatives, is an approximation to the expected Hessian. Furthermore, in each layer $i$ the block $F_i$ can
 be approximated by a Kronecker product of two much smaller matrices $G_i$ and $A_i$, where $G_i = \bld g_i \bld g_i^T$ is the outer product of gradients of pre-activation of $i$-th layer and $A_{i} = \bld a_{i-1}\bld a_{i-1}^T$ is the outer product of activation from the previous layer. 
 This is known as the Kronecker-factored approximate curvature (K-FAC)~\cite{martens2015optimizing}.
%
% By employing Kronecker-product factorization approximate curvature (K-FAC)~\cite{martens2015optimizing}, Fisher information matrix can be approximated by Kronecker product of two much smaller matrices as
 % \begin{equation} \label{laplace form}
 %$\bld{F}_{i} = \bld G_i \otimes \bld A_i$,
 % \end{equation}
 %where $\bld G_i = \bld g_i \bld g_i^T$ is the outer product of gradients of pre-activation of $i$-th layer and $\bld A_{i} = \bld a_{i-1}\bld a_{i-1}^T$ is the outer product of activation from the previous layer, respectively.
 If a Gaussian prior is used and $F$ is scaled by the size of the training set $N$, then the resulting posterior can be written as matrix normal distribution~\cite{gupta1999matrix}:
 \begin{equation}
 \begin{aligned} \label{mvg_prior}
 \bld W_{i} \sim \mathcal{MN}(\bld W_i^\star, (\sqrt{N} \mathbb E[\bld A_i]+\sqrt{\tau}\bld I)^{-1}, (\sqrt{N}\mathbb E[\bld G_i]+\sqrt{\tau}\bld I)^{-1})
 \end{aligned}
 \end{equation} 
 where $\tau$ is the standard deviation of the Gaussian prior.  
In practice, $N$ and $\tau$ can be treated as hyper-parameters as well and tuned on a validation set.

\section{Improvements based on uncertainty estimates}
In this section, we describe how the uncertainty estimates can be utilized with contextual information within \gls{CRF}s for further improvements. 
Then, we introduce how to make use of them in adaptive learning for domain adaptation tasks.

% \subsection{Improving uncertainty estimates with \gls{BNN}}

% We do not insert dropout into the layers before the flatten layer because the pre-trained features would get destroyed and thus leading to a significant drop of performance after fine-tuning.
% Therefore three additional fully connected layers are added to ensure that the model possesses a large enough model capacity. 
% This can reduce the expense of computation and time significantly.
% During training, the two parts are trained together in order to achieve a good balance between the uncertainty estimation and accuracy.
% \begin{figure}[]
%	\begin{center}
%		\includegraphics[height=3.65cm, width=7cm]{m_network}
%		\caption{Modified network architecture of ResNet50.}		
%		\label{fig:modified_net}
%	\end{center}
% \end{figure}

\subsection{Utilizing uncertainty estimates with CRF} \label{combin_crf}
While the BNN approach is very useful in providing reliable uncertainty estimates for single object instances, it does not incorporate any \emph{context information} specific for a scene, such that, e.g. more likely object constellations can be accounted for.
In order to exploit such contextual information within the classification, we combine the output of the \gls{BNN} and the relationships between objects within a scene via a \gls{CRF}(see \fref{fig:combined_crf}).
%As can be seen in \fref{fig:combined_crf}, objects in one scene are firstly fed into \gls{BNN} and their output distributions are fed into the \gls{CRF} as unary features.

In details, we define a scene as a set of $n$ object instances $\mathbf{x}=\{\mathbf{x_1},\dots,\mathbf{x_n}\}$ with corresponding class labels $\mathbf{y}=\{\mathbf{y_1},\dots, \mathbf{y_n}\}$ represented as one-hot encondings, i.e. $\mathbf{y_i} \in \{0,1\}^C$ and $\sum_{j=1}^C y_{ij}=1$, where $C$ is the number of object classes.
The \gls{CRF} models the joint probability $p(\bld y \mid \bld x)$ as an undirected graph consisting of cliques of random variables.
Here a \emph{pairwise} \gls{CRF} is used, consisting of nodes $\mathcal{V}$ and edges $\mathcal{E}$, where the node potentials are modeled as $\phi_{u}(\bld x_i, \bld y_i)$ for individual object instances and the edge potentials $\phi_{p}(\bld x_i, \bld x_j, \bld y_i, \bld y_j)$ for pairs of objects $(\bld x_i, \bld x_j)$ which are in the scene.
Concretely, we define $\phi_{u}$  as the predictive probability of each instance (see Eq.~\eqref{eq:pred_dist}) and $\phi_{p}$ as the co-occurrence probability of two objects.
Co-occurrence probabilities can be obtained from an independent source (as in our household use-case, discussed shortly in \sref{sec:conclusion}).
In case the list of expected objects in the scene is known (as in our industrial use-case, evaluated in \ssref{sec:crf_tless}), the pairwise feature is binary and provided automatically per scene.
Thus, the \gls{CRF} has the following form:

\begin{equation}
\label{crf_used}
p(\bld y|\bld x; \bld \theta) = \frac{1}{Z(\bld x, \bld \theta)}\exp\left(\theta_u\sum_{i\in \mathcal{V}}p(\bld y_i|\bld x_i) + 
\theta_p\sum_{(i,j)\in \mathcal{E}} M (\bld y_i, \bld y_j)\right),
\end{equation}
where $\bld \theta = \{\theta_u, \theta_p\}$ are the node and edge weights respectively, $Z$ is the partition function, and $M$ is a $C\times C$ binary matrix modelling the co-occurrence of two object classes $\bld y_i$ and $\bld y_j$.
The training process of the \gls{CRF} involves minimizing the negative log likelihood, i.e. finding optimal model parameters $\theta^\ast$ such that $\theta^\ast = \arg\min_\theta
\{-\log p(\bld y \mid \bld x; \theta)\}$.
To do this, we employ \gls{SGD} with momentum, which requires the calculation of gradients and thus an inference step for the likelihood shown in Eq.~\eqref{crf_used}.
We use a fully connected \gls{CRF}, i.e. an exact inference of the likelihood is intractable.
Therefore, we apply \gls{LBP} for approximate inference.
In our implementation, we use the C++ library UPGM++~\cite{Ruiz-Sarmiento-REACTS-2015} for this purpose.
%Their contextual relationships in one specific scene are represented by a co-occurrence matrix which act as pairwise features.
%The choice and design of it have large impact on the improvement achieved by \gls{CRF} because the information brought by pairwise features should be complementary rather than contradictory to that of unary features.
%Co-occurrence probabilities can be obtained from an independent source (as in our household use-case, discussed shortly in \sref{sec:conclusion}).
%In case the list of expected objects in the scene is known (as in our industrial use-case, evaluated in \ssref{sec:crf_tless}), the pairwise feature is binary and provided automatically per scene.
% This list of objects can be approximate as well, since only pairwise relations are considered in a weighted smoothing, making its effect less constricting than in an optimal assignment approach for example.

%One thing worth noting, in experiment part, this approach is tested only on the T-LESS dataset, since constructing scenes for objects in the \gls{WRGBD} dataset is hard.
%The reason is that most objects there have not only similar appearances but also similar contexts.
%Therefore it is hard to disambiguate miss-classification by incorporating their contextual information.

\subsection{Adaptive Learning for Domain Adaptation} \label{adp_learning}
The domain gap between the training and test data distribution deteriorates the performance of most of classifiers.
This problem is unavoidable when the classifier is trained on easily obtainable dataset such as a public large-scale or synthetic dataset and then deployed in a real environment.
% because these datasets require less manual labeling efforts, and then deployed in a test environment. 

\begin{figure}[]
	\begin{center}
		\includegraphics[width=12cm,height=6.25cm]{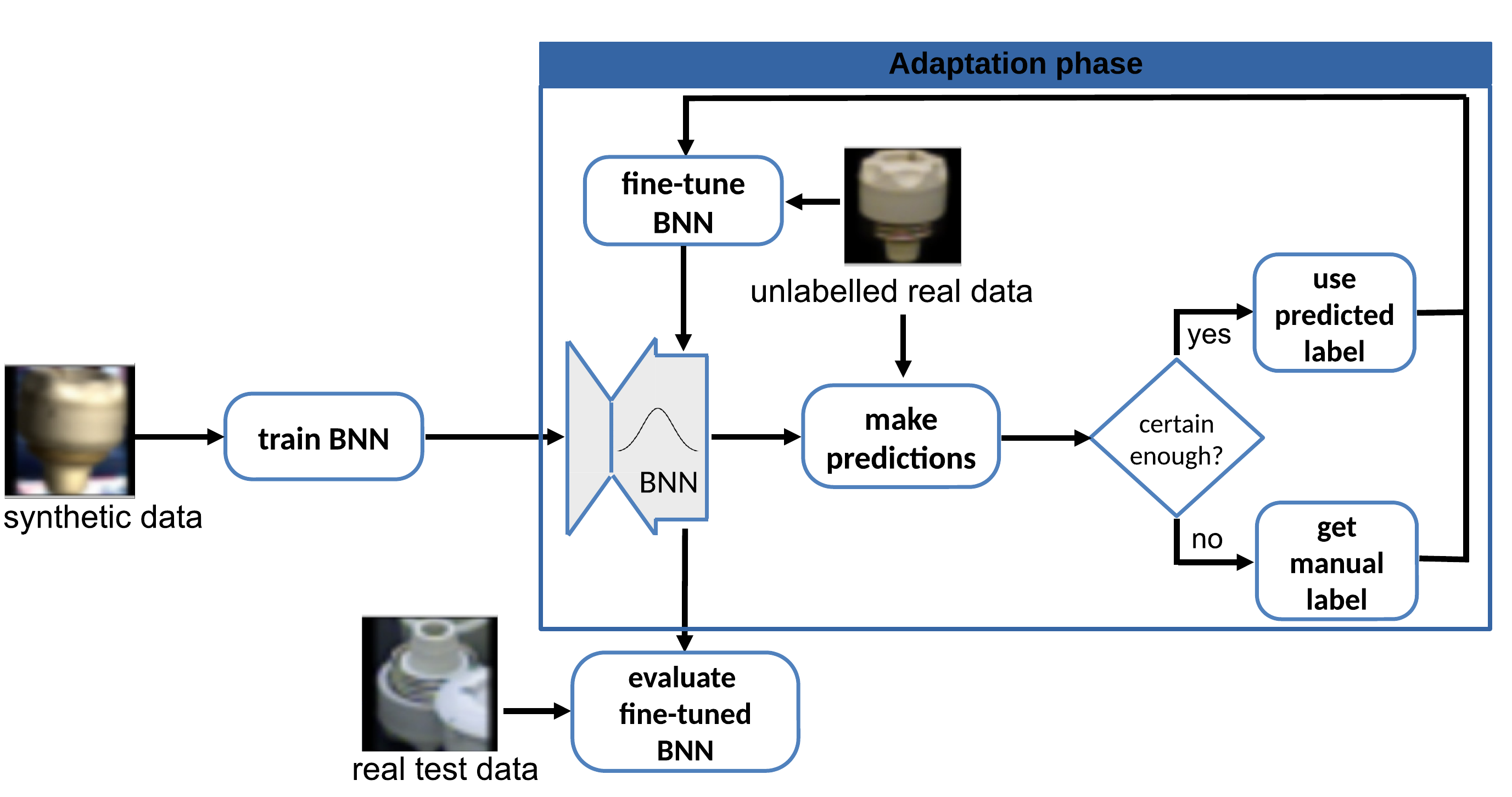}
		\caption{The flowchart for adaptive learning in domain adaptation. Better uncertainty estimates can help distinguishing certain predictions in automatic labeling during adaptation phase (illustrated on the T-LESS dataset and best viewed in color).}		
		\label{fig:adp_learning}
	\end{center}
\end{figure}

In this case, the effects of better uncertainty estimates can be presented by adapting the classifier to the test data with as little manual efforts as possible.
%, in order to resolve the aforementioned problem.
The proposed flowchart for adaptive learning is visualized in \fref{fig:adp_learning}.
For this purpose, the classifier should be introspective, that is, to express reliable confidences about its predictions.
%It is common that the test data in real environment does not have exactly the same distribution as the training set, which leads to a significant performance drop in testing (see arrow A in \fref{fig:adp_learning}). Therefore it would be better to enable the classifier to adapt to the test environment by fine-tuning itself.

At first, the classifier is trained on an easily obtainable or accessible dataset, which can be a large-scale public or synthetic one.
Next, in \textbf{adaptation phase} the classifier is able to adapt to the test data by fine-tuning itself on the so-called adaptation dataset.
In this work, we focus on obtaining this kind of adaptation dataset with as little manual efforts as possible. 
To this end, the annotations in this dataset are collected in a semi-supervised manner (including both automatic and manual manner). 
On the one hand, the predictions with high confidence are used for pseudo labels, thus requiring the classifier to provide reliable uncertainty estimation for both correct and false predictions. 
On the other hand, the classifier would ask people to label a small and random portion of data interactively. 
%The interactive way of labeling is conducted for keeping the adaptation dataset as balanced and diverse as possible.

In the end, the adapted classifier is evaluated on the real test data.
To note that, if the relationships between objects in the test environment are complementary to the \gls{BNN} classifier and can be encoded well with pairwise feature, the \gls{CRF} can be applied to capture them for further improvements. 
% In the experiment part, this approach is evaluated on both UniHB(\gls{WRGBD}) and T-LESS(synthetic T-LESS) data.
% The unadapted classifier can also be tested as a baseline (arrow A).

\section{Experiments}
In this section, we firstly compared performance on uncertainty estimates of two approximate inference techniques for \gls{BNN}s, which are \gls{CDP} and \gls{LAP} on a household objects dataset in terms of comprehensive metrics. 
Then the one with better performance was applied in the following experiments, which are to evaluate \textbf{(1)} the combination with \gls{CRF}s and \textbf{(2)} the adaptive learning for domain adaptation respectively. 

Two types of datasets were employed in our experiments.
The first one is the household objects including the \gls{WRGBD} and the UniHB dataset recorded by ourselves trying to mimic the \gls{WRGBD} but with only one instance in each category. 
They contain multi-view images of household objects in 51 classes, with a $15^{\circ}$ step in elevation (from $30^{\circ}$to $60^{\circ}$) and $2^\circ$ step in azimuth (from $0^\circ$ to $360^\circ$). 
Besides, we have recorded some household objects of novel categories which served as \gls{OOD} dataset. %as well as tabletop scenes captured by our robot.
The second one is an industrial dataset, T-LESS ~\cite{hodan2017tless}, which has little texture but similar appearance between objects. 
This dataset contains multi-view images of industrial components objects in 30 classes. 
The training images depict objects in isolation with a black background, while the test images are from 20 table-top scenes with arbitrarily arranged objects placed on a table (as in a kitting or sorting task).
%In this work only the cropped images of objects in testing scenes are considered, resulting in a testing set with size $\sim$69.5K.
Besides the original T-LESS dataset, we have generated a synthetic dataset trying to mimic the original T-LESS training set.
Since there are lots of occlusions in the test scenes, we employed data augmentations both to the original and synthetic T-LESS training set.  

As mentioned in \ssref{adp_learning}, an easily obtainable dataset is used for training in initialization phase. 
This can be a large-scale public dataset like \gls{WRGBD} dataset or synthetic one like the synthetic T-LESS training set we generated. 
%In adaptation phase, the adaptation dataset requires annotation in both automatic way and interactive way. 
%Therefore uncertainty estimates need to be evaluated on this dataset. 
The (independent) adaptation and testing datasets simulate the data that the classifier encounters in the test environment.
%A reliable uncertainty estimation about the predictions on this dataset can not only assist in collecting data annotations but also avoiding unnecessary accidents.
%We use the objects of $30^{\circ}$ and $60^{\circ}$ in UniHB dataset and the original training set of T-LESS for this purpose.  

\subsection{Uncertainty estimates evaluation}
In this part, we performed extensive experiments to evaluate uncertainty estimates on a household objects dataset. 
We trained models on the entire \gls{WRGBD} dataset and tested them on objects of $30^{\circ}$ and $60^{\circ}$ in the UniHB dataset. 

\begin{figure}[]
	\begin{center}
		\includegraphics[width=\columnwidth,height=7cm]{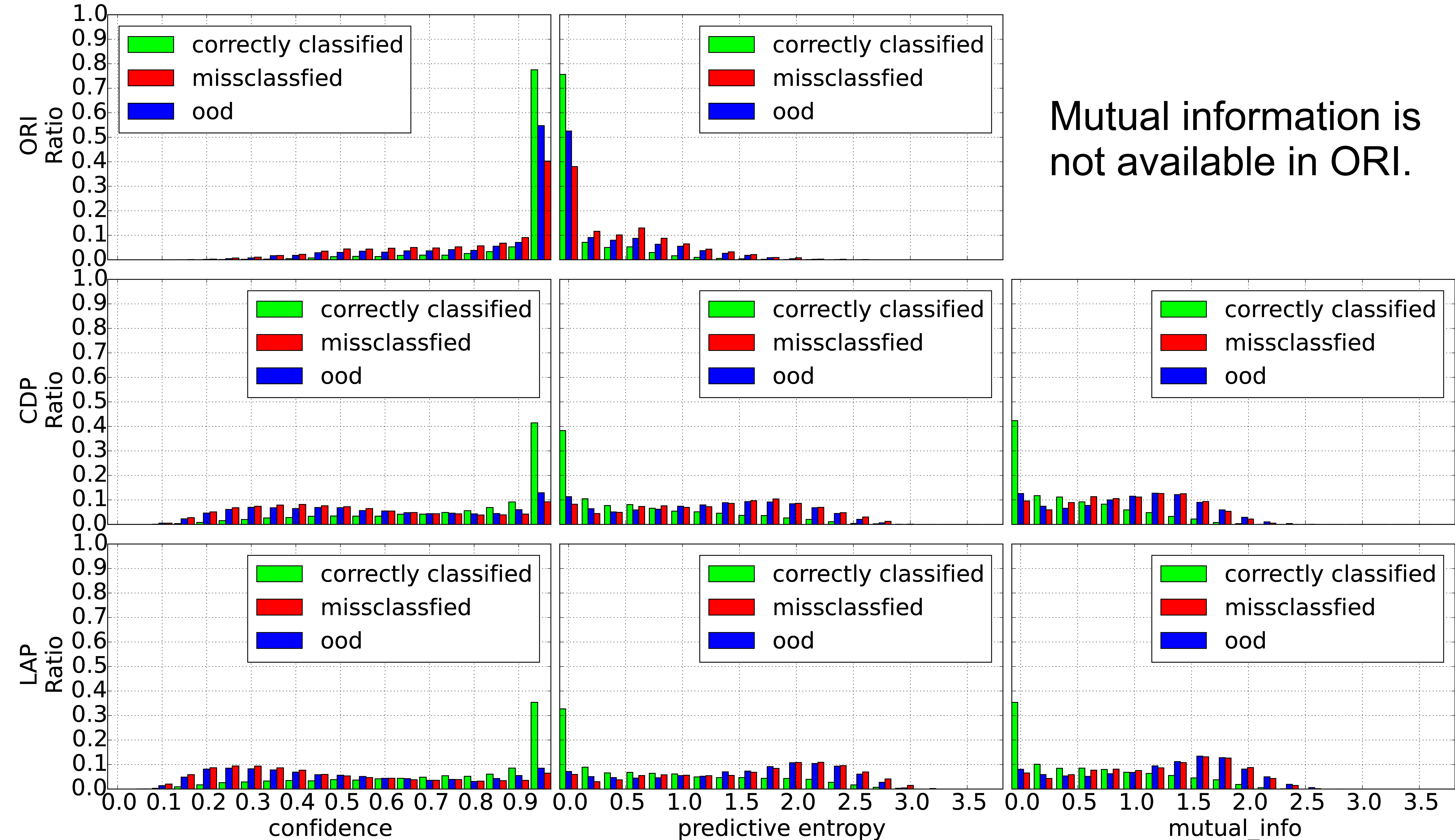}
		\caption{Histograms of three uncertainty measures including confidence, predictive entropy and mutual information of ORI, CDP and LAP in top-down wise (best viewed in color).}		
		\label{fig:uncer_hist}
	\end{center}
\end{figure}

Different metrics were used for the evaluation. 
To evaluate calibration performance we used \gls{ECE} and \gls{MCE}~\cite{guo2017calibration}. 
For summary of both accuracy and calibration we used predictive \gls{NLL} and brier score, which belong to proper scoring rules~\cite{gneiting2007probabilistic}. 
Additionally, we also employed metrics such as area under \gls{ROC} curve and area under \gls{PR} curve to measure the separability between correct predictions and miss-classifications as well as \gls{OOD} predictions. 
Apart from quantitative metrics, a qualitative (visual) metrics, the histogram (see \fref{fig:uncer_hist} \& \fref{fig:crf_hist}) of uncertainty estimates was employed. 
For better visualization, we set the normalizer in the histogram as the amount of the corresponding type of prediction.  
Regarding the uncertainty measure, we evaluated three different ones including confidence (maximum predictive likelihood), predictive entropy and mutual information~\cite{gal2016uncertainty}. The separability metrics list in the \tref{uncer_table} were chosen based on the uncertainty measure with best performance.
 
The \glspl{DNN} and \glspl{BNN} were implemented in Tensorflow %\cite{abadi2016tensorflow},
and the optimization was performed using RMSprop with an initial learning rate of $1e^{-5}$ and L2 regularization with coefficient of $3.5e^{-6}$ as well as the dropout regularization with coefficient of $1.0e^{-5}$.
Early stopping was applied for model selection, %where the amount of epochs to run is determined 
based on the performance on a validation set. 
During inference, the number of samples drawn from the posterior distribution was set to 50 for both inference methods.

In order to preserve the powerful feature extraction capability of ResNet50 and incorporate the better uncertainty estimation from \gls{BNN}s, we slightly modify it by appending three fully connected layers with 1024 hidden units before the output layer.
\gls{CDP}s are inserted into the flatten layer and the three new fully connected layers.
The weights of these layers were initialized from a Gaussian prior ($\mathcal{N}(0, 0.1)$) and the rest from the model pre-trained on ImageNet~\cite{russakovsky2015imagenet}.
This avoids destroying the pre-trained features and enables the model to possess large enough model capacity which was reduced by inserting dropout~\cite{srivastava2014dropout}.
Furthermore, the computation complexity during inference can be reduced by only running the forward pass of the additional layers instead of the whole network.
In the following, we show both qualitative and quantitative results in \fref{fig:uncer_hist} and \tref{uncer_table}, in which we denote original version of ResNet50 by \textbf{ORI} (without additional fully-connected layers), concrete dropout by \textbf{CDP}, Laplace approximation by \textbf{LAP}. 
The point estimate model parameters for \gls{LAP} was model trained with \gls{CDP}. 
We set the hyper-parameter $N$ as 1 and $\tau$ as 15 in \gls{LAP}. 

As can be seen, \gls{BNN}s can achieve better performance of uncertainty estimates in terms of all metrics when compared with ORI. 
At the same time, \gls{CDP} has better performance than \gls{LAP} in terms of proper scoring rules and calibration metrics. 
When \gls{OOD} predictions were considered along with miss-classifications, \gls{ECE} and \gls{MCE} decreased significantly.
This is because prediction of \gls{OOD} data is always incorrect and not all predictions of \gls{OOD} produced high uncertainty correspondingly. If their predictions are highly uncertain, the calibration metrics would have similar values with the ones without \gls{OOD} data.
Both inference methods yield similar results on separability metrics. 
% Regarding to the efficiency of inference, \gls{CDP} can yield acceptable 
% Because the \gls{OOD} data have different appearances and is not present in the training data, it is expected to have larger separability with correct predictions than that between miss-classifications and correct ones. 
Based on these experimental results, we used \gls{CDP} in the following experiments. 

\begin{table}[H]\scriptsize
	\centering
	\caption{Different quantitative results averaged over 3 different random seeds}
	\begin{tabular}{|c|c|c|c|c|c|c|c|}
		\hline
		\label{uncer_table}
		& ACC $\bld \uparrow$                                                    & \begin{tabular}[c]{@{}c@{}}predictive\\ NLL $\bld \downarrow$\end{tabular} & \begin{tabular}[c]{@{}c@{}}Brier\\ score$\bld \downarrow$\end{tabular}  & \begin{tabular}[c]{@{}c@{}}ECE\\ (w/o. OOD/\\  w. OOD)$\bld \downarrow$\end{tabular} & 
		\begin{tabular}[c]{@{}c@{}}MCE\\ (w/o. OOD/\\  w. OOD)$\bld \downarrow$\end{tabular} & 
		\begin{tabular}[c]{@{}c@{}}AUROC\\ (vs. Miss-\\  classified/ \\ vs. OOD)$\bld \uparrow$\end{tabular} & 
		\begin{tabular}[c]{@{}c@{}}AUPR\\ (vs. Miss-\\  classified/ \\ vs. OOD)$\bld \uparrow$\end{tabular} \\ \hline
		ORI & \begin{tabular}[c]{@{}c@{}}0.568\\ $\pm$0.008\end{tabular} & \begin{tabular}[c]{@{}c@{}}3.342\\ $\pm$0.340\end{tabular}   & \begin{tabular}[c]{@{}c@{}}0.722\\ $\pm$0.019\end{tabular} & \begin{tabular}[c]{@{}c@{}}0.304$\pm$0.016/\\ 0.633$\pm$0.065\end{tabular}  & \begin{tabular}[c]{@{}c@{}}0.461$\pm$0.027/\\ 0.362$\pm$0.025\end{tabular}  & \begin{tabular}[c]{@{}c@{}}0.750$\pm$0.007/\\ 0.664$\pm$0.011\end{tabular}                    & \begin{tabular}[c]{@{}c@{}}0.802$\pm$0.008/\\ 0.751$\pm$0.018\end{tabular}                   \\ \hline
		CDP & \begin{tabular}[c]{@{}c@{}}\textbf{0.577}\\ $\pm$0.008\end{tabular} & \begin{tabular}[c]{@{}c@{}}\textbf{2.088}\\ $\pm$0.181\end{tabular}   & \begin{tabular}[c]{@{}c@{}}\textbf{0.594}\\ $\pm$0.013\end{tabular} & \begin{tabular}[c]{@{}c@{}}\textbf{0.124}$\pm$0.023/\\ \textbf{0.288}$\pm$0.048\end{tabular}  & \begin{tabular}[c]{@{}c@{}}\textbf{0.206}$\pm$0.015/\\ \textbf{0.374}$\pm$0.018\end{tabular}  & \begin{tabular}[c]{@{}c@{}}0.775$\pm$0.008/\\ \textbf{0.783}$\pm$0.022\end{tabular}                    & \begin{tabular}[c]{@{}c@{}}0.825$\pm$0.007/\\ \textbf{0.850}$\pm$0.022\end{tabular}                   \\ \hline
		LAP & \begin{tabular}[c]{@{}c@{}}0.576\\ $\pm$0.009\end{tabular} & \begin{tabular}[c]{@{}c@{}}2.322\\ $\pm$0.350\end{tabular}   & \begin{tabular}[c]{@{}c@{}}0.602\\ $\pm$0.011\end{tabular} & \begin{tabular}[c]{@{}c@{}}0.129$\pm$0.058/\\ 0.341$\pm$0.157\end{tabular}  & \begin{tabular}[c]{@{}c@{}}0.235$\pm$0.073/\\ 0.406$\pm$0.070\end{tabular}  & \begin{tabular}[c]{@{}c@{}}\textbf{0.779}$\pm$0.004/\\ 0.782$\pm$0.017\end{tabular}                    & \begin{tabular}[c]{@{}c@{}}\textbf{0.826}$\pm$0.007/\\ 0.849$\pm$0.016\end{tabular}                   \\ \hline
	\end{tabular}
\end{table}

\subsection{Combining with \gls{CRF}s}\label{sec:crf_tless}
In this experiment, we will show the results on evaluating the idea introduced in \ssref{combin_crf}. 
%To evaluate this idea quickly, we firstly train \gls{DNN} and \gls{BNN} with concrete dropout on the original T-LESS training set.
%There are 20 scenes in the T-LESS test set. 
%In order to evaluate this idea quickly and in order to obtain co-occurrence information,
We use the test set of T-LESS in this part.
We split the scenes 2, 3, 5, 8 off for training our \gls{CRF} and the scenes 1, 4, 6, 7 for testing. 
%Since co-occurrences can only be obtained by scenes we split the 20 test scenes for this experiment in: 2, 3, 5, 8 for training the \gls{CRF} and 1, 4, 6, 7 for testing.
%The reason for this kind of splitting is to keep as many categories as possible occurring in both sets, training and testing.
These splits were chosen in this way so that as many categories as possible occur in both training and testing (an evaluation on the whole T-LESS test set is shown in the next experiment).
The maximum number of iterations during training is 30K, the initial learning rate is $1e^{-4}$, and the size of mini-batch is 16.

%\com{revise this part if it is correct and understandable; I based this part on JX comment: "In this context, DNN means network trained without dropout, NOMCD means network trained with dropout but testing with MCD off, BNN means network trained with dropout but testing with MCD on." I hope I understood it correctly :P}
In order to see the influence of reliable uncertainty estimates we firstly trained \gls{DNN}s and \gls{BNN}s which provide the unary potential in the next step.
%\todo{introducing MCD here without any explanation/introduction before is a bit confusing. What do you think?}
Preliminary experiments, which are not displayed here, show a significant lower performance of the \gls{DNN} trained without dropout compared to the \gls{BNN}.
On the other hand, the \gls{DNN} trained with dropout but turning off \gls{MCD} during inference (denoted as \textbf{NOMCD} in the following) resulted in worse uncertainty estimates but a better accuracy.
Hence, since we want to investigate the effect of uncertainty estimates on the \gls{CRF}s, we compared the proposed \gls{BNN} with the \textbf{NOMCD}.
Comparing the weights obtained by training the \gls{CRF} with the uncertainty estimates of NOMCD ($\theta_u$=~4.875; $\theta_p$=~6.073) and \gls{BNN} ($\theta_u$=~8.122; $\theta_p$=~6.59) a different rating of the provided information can be observed ($\theta_u$ vs. $\theta_p$).
While in the \gls{BNN} case the \gls{CRF} relies more on the classifier, in the NOMCD case the co-occurrence statistics are given a higher importance,
reflecting the added usefulness of the correct uncertainty estimates (since the NOMCD and BNN accuracies without smoothing are similar, as seen in \tref{table:crf_exp1_acc}).

\begin{table}[]\footnotesize
	\centering
	\caption{Results of \gls{CRF} trained and tested with different unary features}
	\begin{tabular}{|c|c|c|c|}
		\hline
		\label{table:crf_exp1_acc}
		& \begin{tabular}[c]{@{}c@{}}type of unary features\\ in testing\end{tabular} & \begin{tabular}[c]{@{}c@{}}accuracy with \\ unary potentials\end{tabular} & \begin{tabular}[c]{@{}c@{}}accuracy with \\ unary and pairwise\\ potentials\end{tabular} \\ \hline
		\multirow{2}{*}{\begin{tabular}[c]{@{}c@{}}CRF trained with unary \\  features  from NOMCD\end{tabular}} & NOMCD                                                                         & 58.48\%                                                              & 68.6\%                                                                             \\ \cline{2-4} 
		& BNN                                                                         & 60.36\%                                                              & 76.19\%                                                                             \\ \hline
		\multirow{2}{*}{\begin{tabular}[c]{@{}c@{}}CRF trained with unary \\ features  from BNN\end{tabular}} & NOMCD                                                                         & 58.48\%                                                              & 68.62\%                                                                                \\ \cline{2-4} 
		& BNN                                                                         & 60.36\%                                                              & \textbf{76.36}\%                                                                             \\ \hline
	\end{tabular}
\end{table}

% \begin{table}[]
%	\centering
%	\caption{Weights of CRF trained with different unary features}
%	\begin{tabular}{|c|c|c|}
%		\hline
%		\multicolumn{1}{|l|}{}                                            & \begin{tabular}[c]{@{}c@{}}node weight\\ $\theta_u$ \end{tabular} & \begin{tabular}[c]{@{}c@{}}edge weight\\ $\theta_p$ \end{tabular} \\ \hline
%		\begin{tabular}[c]{@{}c@{}}CRF trained with\\ unary features from ORI\end{tabular} & 3.85                                                   & 4.59                                                   \\ \hline
%		\begin{tabular}[c]{@{}c@{}}CRF trained with\\ unary features from BNN\end{tabular} & 4.94                                                   & 
%		4.23                                                   \\ \hline
%	\end{tabular}
%	\label{table:crf_exp1_weights}
%\end{table}

\begin{figure}[]
	\begin{center}
		\includegraphics[width=\columnwidth]{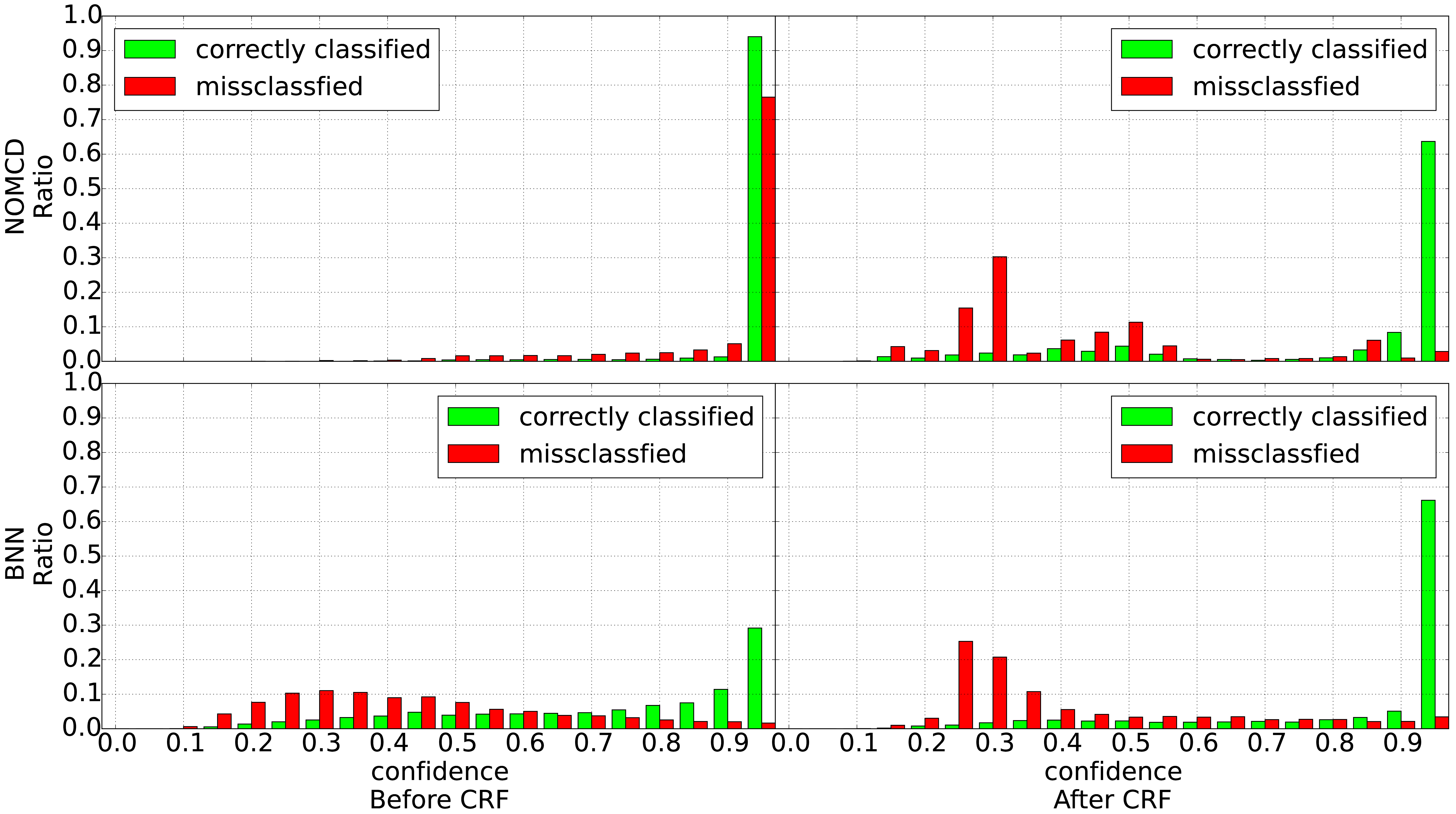}
		\caption{Histograms of confidence of NOMCD (top row) and \gls{BNN} (bottom row) before (left column) and after (right column) applying \gls{LBP} in \gls{CRF} (best viewed in color).}		
		\label{fig:crf_hist}
	\end{center}
\end{figure}

\tref{table:crf_exp1_acc} shows the much larger performance gain when using the \gls{CRF}s with better uncertainty estimates, and this is irrespective of the CRF weights used. %(when testing both network with the averaged $\theta_u$ and $\theta_p$ values, the accuracy also only changes minimally).
%\com{imo the anova results and the explanation of table 2 should come after the explanation of the weights which means here :)}
Besides the performance gain, the \gls{CRF} is also improving (or at least maintaining) the uncertainty estimates. 
\fref{fig:crf_hist} shows the histogram of confidence of the predictions made by NOMCD and \gls{BNN} before and after applying \gls{LBP} inference within \gls{CRF}.
We can see that the uncertainty estimates' quality of NOMCD has been improved and that of \gls{BNN} has been maintained, which can be helpful for further improvement in the down-stream tasks. 

\subsection{Adaptive Learning}
In this part, a proof-of-concept experiment is performed to evaluate the idea illustrated in \sref{adp_learning}. 
To this end, we employed both datasets from two different scenarios for evaluation. 

Following the pipeline in \fref{fig:adp_learning}, at the beginning we used \gls{WRGBD} dataset and the augmented, synthetic T-LESS dataset generated by ourselves for initial training, because they can be obtained more easily.
During adaptation phase, objects of $30^{\circ}$ and $60^{\circ}$ in UniHB dataset ($\sim$17.1K) and the original training set of T-LESS ($\sim$30K) were used as adaptation dataset.
In order to adapt to the test environment, the classifier should be able to collect a dataset for fine-tuning with as little manual efforts as possible.
Therefore, this collected dataset can be annotated in two different manners, automatically and manually.
The automatically labeled data was selected based on threshold of the uncertainty estimates. %(which was chosen based on the validation set).
%, which is chosen based on threshold up to which the accuracy reaches 95\% on an separate validation set.
In the end, during the deployment phase, the adapted model was evaluated on the test dataset. 
The $45^{\circ}$ objects in UniHB dataset and original test set of T-LESS were treated as data the robot encounters in the test environment.
% 
% TODO: should be based on manually labeleled data!
% The threshold is chosen based on accuracy on a small validation set and applied to the whole adaptation dataset.
% In the following, we firstly show the results on household dataset and then on industrial components dataset.

\paragraph{Household objects dataset:}
We tested different versions of the proposed automatic labeling procedure based on uncertainty estimates, and found that the best results were obtained by setting the confidence threshold s.t. the accuracy of the predictions (estimated on a small manually labeled set) is 95\%.
The accuracy of automatically labeled data in \RNum{3}, \RNum{4} is around 96\%, matching the 95\% estimate.
% It is assumed that all the predictions except for automatic labeling have been manually and interactively labeled.

Our main results are shown in \tref{table:fine_tuning}.
As it can be seen, the manual labeling effort can be reduced based on automatic labeling.  
More detailed testing will be performed on the industrial dataset, based on the insights gained here.

During the experiment, we found that the balance of number of each class on the adaptation dataset plays an important role. 
The main reason for this should be the different visual domain gap of different objects. 
The initial model is more familiar with some objects instead of other and thus give lower uncertainty for these familiar ones. 
Since we selected predictions based on the uncertainty estimates, this would lead to an imbalanced dataset and thus bias the adapted model.
Therefore it's important to mitigate this issue.
We found that adding manually labeled data and augmentations is useful not only to increase the diversity of the dataset, but to balance the dataset (see \RNum{3} and \RNum{4}).
Other ways of balancing the automatically labeled data (e.g. by selecting the top most confident predictions per class) decreased performance as they resulted in either too few labels or included too many incorrect ones.

\begin{table}[]\footnotesize
	\centering
	\caption{Results of fine-tuned network on household objects dataset}
	\begin{tabular}{|l|c|}
		\hline
		Dataset used for fine-tuning                                                                                                                                                                                   & \multicolumn{1}{l|}{\begin{tabular}[c]{@{}l@{}}Accuracy (average \\ over 3 random seeds)\end{tabular}} \\ \hline
		\RNum{1}: 0\% (no fine-tuning)                                                                                                                                                                                                  & 66.9\%                                                                                                  \\ \hline
		\begin{tabular}[c]{@{}l@{}}\RNum{2}: 3\% manually labeled data, selected randomly (\textbf{balanced})\end{tabular}                                                                                                                               & 91.7\%                                                                                                  \\ \hline
		\begin{tabular}[c]{@{}l@{}}\RNum{3}: 3\% automatically labeled data (\textbf{imbalanced})\end{tabular}                                                                                                                        & 79.0\%                                                                                                  \\ \hline
		\begin{tabular}[c]{@{}l@{}}\RNum{4}: 2\% automatically labeled data and 1\% manually labeled \\ data randomly, augmentation for balance (\textbf{balanced})\end{tabular}                                                            & 89.6\%                                                                                                  \\ \hline
	\end{tabular}
	\label{table:fine_tuning}
\end{table}

\paragraph{Industrial components dataset:}
With the same procedure of selecting automatically labeled data, the size of dataset is $\sim$1K with only 93\% accuracy using the original ResNet50, but $\sim$1.6K with 96\% accuracy using \gls{BNN}.
The summary of the results is shown in \tref{table:fine_tuning_TLESS}.
The performance of the classifier adapted using 3\% manually labeled data (\RNum{6}) is matched by the use of 1\% manually labeled data if automatic labeling is employed (\RNum{5}). Moreover, adding the automatic labeling to the 3\% manually labeled data can nearly reach the the performance of classifier adapted with all available data manually labeled (\RNum{3} vs \RNum{7}). By incorporating contextual information with \gls{CRF}, the performance can be increased further (\RNum{8}).

\begin{table}[]\footnotesize
	\centering
	\caption{\gls{BNN} fine-tuning with different datasets (size of dataset before augmentations is showed in the bracket).}
	\begin{tabular}{|l|c|}
		\hline
		Dataset used for fine-tuning                                                                                                                                    & \multicolumn{1}{l|}{Accuracy} \\ \hline
		\RNum{1}: augmented, synthetic dataset & 34.91\% \\ \hline
		\RNum{2}: fine-tune \RNum{1} with augmented, automatically labeled real dataset ($\sim$1.6K) & 53.54\% \\ \hline
		\RNum{3}: entire real dataset, i.e. 100\% maually labeled ($\sim$30K), augmented & 72.78\% \\ \hline
		\RNum{4}: fine-tune \RNum{1} with 1\% manually labeled real dataset ($\sim$0.3K), augmented & 67.4\% \\ \hline
% 		\begin{tabular}[c]{@{}l@{}}\RNum{5}: fine-tune \RNum{1} with automatically labeled \\ and 1\% manually labeled real dataset ($\sim$1.9K), augmented\end{tabular} & 68.1\% \\ \hline
		\RNum{5}: fine-tune \RNum{1} with \RNum{2} and \RNum{4} ($\sim$1.9K), augmented & 68.1\% \\ \hline
		\RNum{6}: fine-tune \RNum{1} with 3\% manually labeled real dataset ($\sim$0.9K), augmented & 68.1\% \\ \hline
% 		\begin{tabular}[c]{@{}l@{}}\RNum{7}: fine-tune \RNum{1} with automatically labeled \\ and 3\% manually labeled real dataset ($\sim$2.5K), augmented\end{tabular} & 72.48\% \\ \hline
		\RNum{7}: fine-tune \RNum{1} with \RNum{2} and \RNum{6} ($\sim$2.5K), augmented & 72.48\% \\ \hline
		\RNum{8}: Incorporating contextual information with \gls{CRF} based on \RNum{7} & 74.64\% \\ \hline
	\end{tabular}
	\label{table:fine_tuning_TLESS}
\end{table}

\section{Conclusions}
\label{sec:conclusion}
We presented an approach to make robots learning new objects more introspectively, by improving its awareness of possible mistakes, and leveraging this in two ways: first, for better incorporating context information (if available) through smoothing over all object predictions using a CRF, and second, for exploiting this in semi-supervised domain adaptation, where the mostly correct predictions are automatically obtained as adaptation data while asking humans for help with the more uncertain ones.

The improved uncertainty estimation from \gls{BNN} plays an important role especially in the latter use-case, because it not only provides a reliable uncertainty estimation, but also increases the separability between correct predictions and false predictions, which is more useful in this task. 
It was found, however that it is very important to ensure that the data is balanced.
For manual labeling this can be easily achieved by requesting the human operator to label a more-or-less equal number of instances of each object, e.g. repeatedly selecting random subsets and having to click all occurrences of an object (as in an image CAPTCHA), then switching to the next target object once enough samples were collected.
For the automatic labeling, random selection is not a good alternative, as the accuracy penalty would be too large if the overall performance of the initial classifier is too low (as in our cases). It could be, however, incorporated if multiple rounds of adaptation are performed, and the performance is gradually increasing to acceptable levels (around 95\% in our tests).

In the former use-case the importance of a clear co-occurence statistic is highlighted by the fact that the CRF failed to improve results on the household dataset (the pairwise weight was negligible) due to the difficulty of obtaining good co-occurrence statistics in this scenario (which we mined from word co-occurences in WikiHow articles) and since many household objects have similar appearances and contexts at the same time.
In an industrial scenario, e.g. for kitting applications, such a list of parts is available, and the learned CRF weights generalize well over objects and scenes.
 
\textbf{Acknowledgments:} This work was partially funded by the Big Data Interdisciplinary Project of DLR e.V. under the project number 2464047. Jianxiang Feng is supported by the Munich School for Data Science (MUDS) and Rudolph Triebel is a member of MUDS.

%Our next step will be to incorporate the proposed approach into detectors.

% \section{Acknowledgments}
% \todo{students?}

%
% ---- Bibliography ----
%
\bibliographystyle{styles/bibtex/splncs03_unsrt}
\bibliography{references}

\begin{thebibliography}{10}
\providecommand{\url}[1]{\texttt{#1}}
\providecommand{\urlprefix}{URL }

\bibitem{gal2017deep}
Gal, Y., Islam, R., Ghahramani, Z.: Deep bayesian active learning with image
  data. In: Proceedings of the 34th International Conference on Machine
  Learning-Volume 70. pp. 1183--1192. JMLR. org (2017)

\bibitem{blundell2015weight}
Blundell, C., Cornebise, J., Kavukcuoglu, K., Wierstra, D.: Weight uncertainty
  in neural network. In: International Conference on Machine Learning. pp.
  1613--1622 (2015)

\bibitem{osband2016deep}
Osband, I., Blundell, C., Pritzel, A., Van~Roy, B.: Deep exploration via
  bootstrapped dqn. In: Advances in neural information processing systems. pp.
  4026--4034 (2016)

\bibitem{gal2016improving}
Gal, Y., McAllister, R., Rasmussen, C.E.: Improving pilco with bayesian neural
  network dynamics models. In: Data-Efficient Machine Learning workshop, ICML
  (2016)

\bibitem{grimmett15introspective}
Grimmett, H., Triebel, R., Paul, R., Posner, I.: Introspective classification
  for robot perception. The International Journal of Robotics Research (IJRR)
  35(7),  743--762 (2016)

\bibitem{hendrycks2016baseline}
Hendrycks, D., Gimpel, K.: A baseline for detecting misclassified and
  out-of-distribution examples in neural networks. In: 5th International
  Conference on Learning Representations, {ICLR} 2017, Toulon, France, April
  24-26, 2017, Conference Track Proceedings (2017),
  \url{https://openreview.net/forum?id=Hkg4TI9xl}

\bibitem{kurakin2016adversarial}
Kurakin, A., Goodfellow, I., Bengio, S.: Adversarial machine learning at scale.
  arXiv preprint arXiv:1611.01236  (2016)

\bibitem{balan2015bayesian}
Balan, A.K., Rathod, V., Murphy, K.P., Welling, M.: Bayesian dark knowledge.
  In: Advances in Neural Information Processing Systems. pp. 3438--3446 (2015)

\bibitem{gal2016dropout}
Gal, Y., Ghahramani, Z.: Dropout as a bayesian approximation: Representing
  model uncertainty in deep learning. In: international conference on machine
  learning. pp. 1050--1059 (2016)

\bibitem{louizos2016structured}
Louizos, C., Welling, M.: Structured and efficient variational deep learning
  with matrix gaussian posteriors. In: International Conference on Machine
  Learning. pp. 1708--1716 (2016)

\bibitem{gal2017concrete}
Gal, Y., Hron, J., Kendall, A.: Concrete dropout. In: Advances in Neural
  Information Processing Systems. pp. 3581--3590 (2017)

\bibitem{sun2017learning}
Sun, S., Chen, C., Carin, L.: Learning structured weight uncertainty in
  bayesian neural networks. In: Artificial Intelligence and Statistics. pp.
  1283--1292 (2017)

\bibitem{louizos2017multiplicative}
Louizos, C., Welling, M.: Multiplicative normalizing flows for variational
  bayesian neural networks. In: Proceedings of the 34th International
  Conference on Machine Learning. vol.~70, pp. 2218--2227. JMLR. org (2017)

\bibitem{ritter2018scalable}
Ritter, H., Botev, A., Barber, D.: A scalable laplace approximation for neural
  networks. In: International Conference on Learning Representations (2018),
  \url{https://openreview.net/forum?id=Skdvd2xAZ}

\bibitem{wang2018adversarial}
Wang, K., Vicol, P., Lucas, J., Gu, L., Grosse, R.B., Zemel, R.S.: Adversarial
  distillation of bayesian neural network posteriors. In: Proceedings of the
  35th International Conference on Machine Learning, {ICML} 2018,
  Stockholmsm{\"{a}}ssan, Stockholm, Sweden, July 10-15, 2018. pp. 5177--5186
  (2018)

\bibitem{lakshminarayanan2017simple}
Lakshminarayanan, B., Pritzel, A., Blundell, C.: Simple and scalable predictive
  uncertainty estimation using deep ensembles. In: Advances in Neural
  Information Processing Systems. pp. 6402--6413 (2017)

\bibitem{koller2009probabilistic}
Koller, D., Friedman, N.: Probabilistic graphical models: principles and
  techniques. MIR Press (2009)

\bibitem{tompson2014joint}
Tompson, J.J., Jain, A., LeCun, Y., Bregler, C.: Joint training of a
  convolutional network and a graphical model for human pose estimation. In:
  Advances in neural information processing systems. pp. 1799--1807 (2014)

\bibitem{liu2015deep}
Liu, F., Shen, C., Lin, G.: Deep convolutional neural fields for depth
  estimation from a single image. In: Proceedings of the IEEE Conference on
  Computer Vision and Pattern Recognition. pp. 5162--5170 (2015)

\bibitem{wang2016towards}
Wang, H., Yeung, D.Y.: Towards bayesian deep learning: A survey. arXiv preprint
  arXiv:1604.01662  (2016)

\bibitem{johnson2016composing}
Johnson, M., Duvenaud, D.K., Wiltschko, A., Adams, R.P., Datta, S.R.: Composing
  graphical models with neural networks for structured representations and fast
  inference. In: Advances in neural information processing systems. pp.
  2946--2954 (2016)

\bibitem{kolesnikov2019revisiting}
Kolesnikov, A., Zhai, X., Beyer, L.: Revisiting self-supervised visual
  representation learning. arXiv preprint arXiv:1901.09005  (2019)

\bibitem{tang2012shifting}
Tang, K., Ramanathan, V., Fei-Fei, L., Koller, D.: Shifting weights: Adapting
  object detectors from image to video. In: Advances in Neural Information
  Processing Systems. pp. 638--646 (2012)

\bibitem{zou2018unsupervised}
Zou, Y., Yu, Z., Vijaya~Kumar, B., Wang, J.: Unsupervised domain adaptation for
  semantic segmentation via class-balanced self-training. In: Proceedings of
  the European Conference on Computer Vision (ECCV). pp. 289--305 (2018)

\bibitem{xu2019self}
Xu, J., Xiao, L., Lopez, A.M.: Self-supervised domain adaptation for computer
  vision tasks. arXiv preprint arXiv:1907.10915  (2019)

\bibitem{wang2016cost}
Wang, K., Zhang, D., Li, Y., Zhang, R., Lin, L.: Cost-effective active learning
  for deep image classification. IEEE Transactions on Circuits and Systems for
  Video Technology  27(12),  2591--2600 (2016)

\bibitem{lin2017active}
Lin, L., Wang, K., Meng, D., Zuo, W., Zhang, L.: Active self-paced learning for
  cost-effective and progressive face identification. IEEE transactions on
  pattern analysis and machine intelligence  40(1),  7--19 (2017)

\bibitem{mackay1992practical}
MacKay, D.J.: A practical bayesian framework for backpropagation networks.
  Neural computation  4(3),  448--472 (1992)

\bibitem{neal2012bayesian}
Neal, R.M.: Bayesian learning for neural networks, vol. 118. Springer Science
  \& Business Media (2012)

\bibitem{graves2011practical}
Graves, A.: Practical variational inference for neural networks. In: Advances
  in neural information processing systems. pp. 2348--2356 (2011)

\bibitem{liu2015crf}
Liu, F., Lin, G., Shen, C.: Crf learning with cnn features for image
  segmentation. Pattern Recognition  48(10),  2983--2992 (2015)

\bibitem{srivastava2014dropout}
Srivastava, N., Hinton, G., Krizhevsky, A., Sutskever, I., Salakhutdinov, R.:
  Dropout: a simple way to prevent neural networks from overfitting. The
  Journal of Machine Learning Research  15(1),  1929--1958 (2014)

\bibitem{gal2016uncertainty}
Gal, Y.: Uncertainty in deep learning. Ph.D. thesis, PhD thesis, University of
  Cambridge (2016)

\bibitem{maddison2016concrete}
Maddison, C.J., Mnih, A., Teh, Y.W.: The concrete distribution: A continuous
  relaxation of discrete random variables. arXiv preprint arXiv:1611.00712
  (2016)

\bibitem{martens2015optimizing}
Martens, J., Grosse, R.: Optimizing neural networks with kronecker-factored
  approximate curvature. In: International conference on machine learning. pp.
  2408--2417 (2015)

\bibitem{gupta1999matrix}
Gupta, A., Nagar, D.: Matrix Variate Distributions, vol. 104. CRC Press (1999)

\bibitem{Ruiz-Sarmiento-REACTS-2015}
Ruiz-Sarmiento, J.R., Galindo, C., Gonz{\'{a}}lez-Jim{\'{e}}nez, J.: Upgmpp: a
  software library for contextual object recognition. In: 3rd. Workshop on
  Recognition and Action for Scene Understanding (REACTS) (2015)

\bibitem{lai2011large}
Lai, K., Bo, L., Ren, X., Fox, D.: A large-scale hierarchical multi-view rgb-d
  object dataset. In: 2011 IEEE international conference on robotics and
  automation. pp. 1817--1824. IEEE (2011)

\bibitem{hodan2017tless}
Hoda{\v{n}}, T., Haluza, P., Obdr{\v{z}}{\'a}lek, {\v{S}}., Matas, J.,
  Lourakis, M., Zabulis, X.: {T-LESS}: An {RGB-D} dataset for {6D} pose
  estimation of texture-less objects. IEEE Winter Conference on Applications of
  Computer Vision (WACV)  (2017)

\bibitem{guo2017calibration}
Guo, C., Pleiss, G., Sun, Y., Weinberger, K.Q.: On calibration of modern neural
  networks. In: Proceedings of the 34th International Conference on Machine
  Learning-Volume 70. pp. 1321--1330. JMLR. org (2017)

\bibitem{gneiting2007probabilistic}
Gneiting, T., Balabdaoui, F., Raftery, A.E.: Probabilistic forecasts,
  calibration and sharpness. Journal of the Royal Statistical Society: Series B
  (Statistical Methodology)  69(2),  243--268 (2007)

\bibitem{russakovsky2015imagenet}
Russakovsky, O., Deng, J., Su, H., Krause, J., Satheesh, S., Ma, S., Huang, Z.,
  Karpathy, A., Khosla, A., Bernstein, M., et~al.: Imagenet large scale visual
  recognition challenge. International journal of computer vision  115(3),
  211--252 (2015)

\end{thebibliography}
% \begin{thebibliography}{6}
% %
% 
% \bibitem {smit:wat}
% Smith, T.F., Waterman, M.S.: Identification of common molecular subsequences.
% J. Mol. Biol. 147, 195?197 (1981). \url{doi:10.1016/0022-2836(81)90087-5}
% 
% \bibitem {may:ehr:stein}
% May, P., Ehrlich, H.-C., Steinke, T.: ZIB structure prediction pipeline:
% composing a complex biological workflow through web services.
% In: Nagel, W.E., Walter, W.V., Lehner, W. (eds.) Euro-Par 2006.
% LNCS, vol. 4128, pp. 1148?1158. Springer, Heidelberg (2006).
% \url{doi:10.1007/11823285_121}
% 
% \bibitem {fost:kes}
% Foster, I., Kesselman, C.: The Grid: Blueprint for a New Computing Infrastructure.
% Morgan Kaufmann, San Francisco (1999)
% 
% \bibitem {czaj:fitz}
% Czajkowski, K., Fitzgerald, S., Foster, I., Kesselman, C.: Grid information services
% for distributed resource sharing. In: 10th IEEE International Symposium
% on High Performance Distributed Computing, pp. 181?184. IEEE Press, New York (2001).
% \url{doi: 10.1109/HPDC.2001.945188}
% 
% \bibitem {fo:kes:nic:tue}
% Foster, I., Kesselman, C., Nick, J., Tuecke, S.: The physiology of the grid: an open grid services architecture for distributed systems integration. Technical report, Global Grid
% Forum (2002)
% 
% \bibitem {onlyurl}
% National Center for Biotechnology Information. \url{http://www.ncbi.nlm.nih.gov}
% 
% 
% \end{thebibliography}
\end{document}